\documentclass[10pt,twocolumn,letterpaper]{article}

\usepackage{cvpr}              
\definecolor{cvprblue}{rgb}{0.21,0.49,0.74}
\usepackage[pagebackref,breaklinks,colorlinks,allcolors=cvprblue]{hyperref}

\usepackage{graphicx}
\usepackage{amsmath}
\usepackage{amssymb}
\usepackage{booktabs}
\usepackage{multicol}
\usepackage{multirow}
\usepackage{soul}
\usepackage{threeparttable} 
\usepackage{colortbl}
\usepackage{tcolorbox}
\usepackage{pifont}
\usepackage[capitalize]{cleveref}
\usepackage{mdframed}

\usepackage[ruled,vlined]{algorithm2e}
\definecolor{DarkBlue}{RGB}{64,101,149}
\definecolor{myyellow}{RGB}{252, 240, 199}

\makeatletter
\newcommand{\thickhline}{%
    \noalign {\ifnum 0=`}\fi \hrule height 0.7pt
    \futurelet \reserved@a \@xhline
}
\tcbuselibrary{listingsutf8}
\newtcbox{\tcbhighmath}[1][]{on line, colframe=white,
  colback=yellow!20, boxrule=0pt, arc=3pt, boxsep=0pt, left=3pt, right=3pt, top=3pt, bottom=3pt, #1}

\newcommand{\ourmethod}{{\fontfamily{lmtt}\selectfont \textbf{EMO-R3}}\xspace}

\def\paperID{10848} 
\def\confName{CVPR}
\def\confYear{2026}

\title{\raisebox{-0.2\height}{\includegraphics[width=0.07\textwidth]{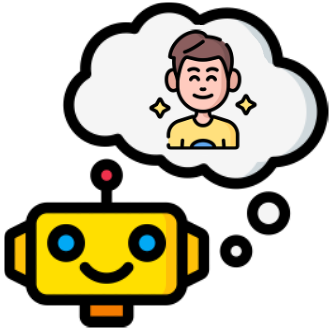}} EMO-R3: Reflective Reinforcement Learning for Emotional Reasoning in Multimodal Large Language Models}

\author{
Yiyang Fang$^{12}$, Wenke Huang$^{1}$, Pei Fu$^{2}$\thanks{Project Leader. Work done during internship at Xiaomi Inc.}\hspace{2pt}, Yihao Yang$^{1}$, Kehua Su$^{1}$, Zhenbo Luo$^{2}$, Jian Luan$^{2}$, Mang Ye$^{1}$\thanks{Corresponding Author.}
\\ 
$^{1}$ School of Computer Science, Wuhan University.\\
$^{2}$ MiLM Plus, Xiaomi Inc.\\
\tt{\small{\{fangyiyang, yemang\}@whu.edu.cn}}\\
{\tt{\small{\url{https://github.com/SeerRay-Lab/emo-r3}}}}
}

\begin{document}
\maketitle
\begin{abstract}
Multimodal Large Language Models (MLLMs) have shown remarkable progress in visual reasoning and understanding tasks but still struggle to capture the complexity and subjectivity of human emotions. Existing approaches based on supervised fine-tuning often suffer from limited generalization and poor interpretability, while reinforcement learning methods such as Group Relative Policy Optimization fail to align with the intrinsic characteristics of emotional cognition.
To address these challenges, we propose Reflective Reinforcement Learning for Emotional Reasoning (\ourmethod{}), a framework designed to enhance the emotional reasoning ability of MLLMs. Specifically, we introduce Structured Emotional Thinking to guide the model to perform step-by-step emotional reasoning in a structured and interpretable manner, and design a Reflective Emotional Reward that enables the model to re-evaluate its reasoning based on visual-text consistency and emotional coherence. Extensive experiments demonstrate that \ourmethod{} significantly improves both the interpretability and emotional intelligence of MLLMs, achieving superior performance across multiple visual emotional understanding benchmarks.
\end{abstract}    
\begin{figure}[htb]
    \centering
	\includegraphics[width=0.96\linewidth]{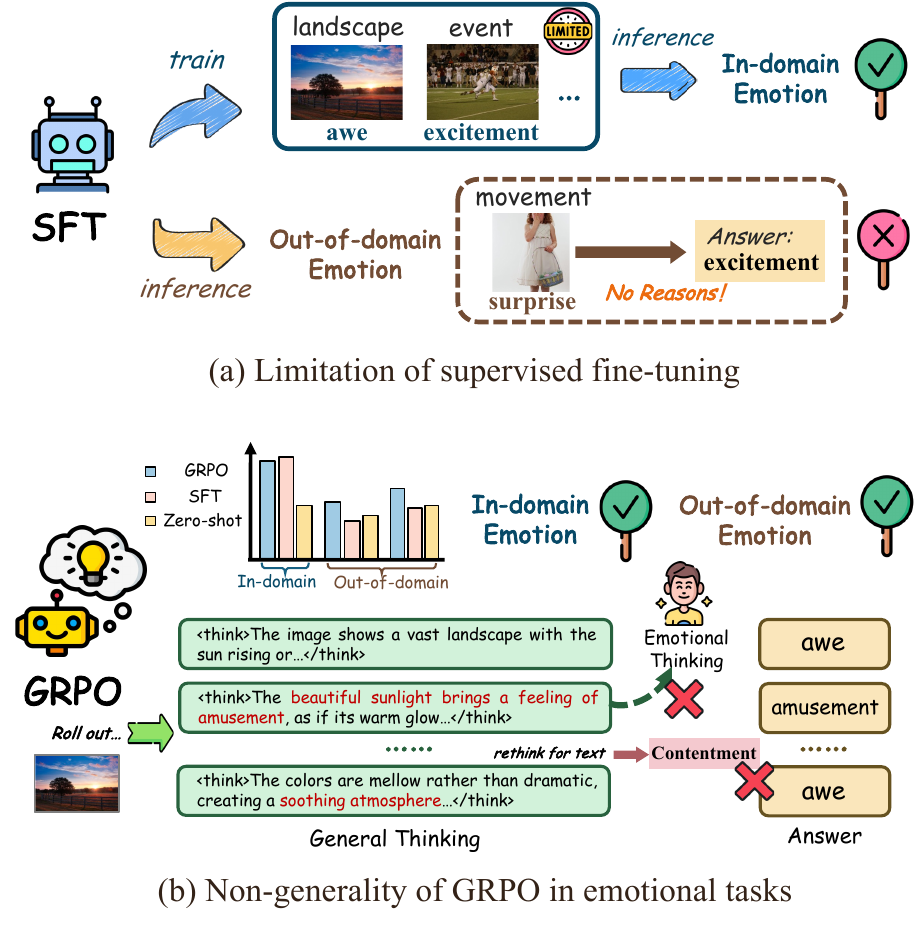}
    \vspace{-0.3cm}
    \caption{
    	\textbf{Illustration of the motivation.}
        \textbf{(a)} SFT relies on human annotations but is constrained by fixed labels and limited categories, resulting in poor generalization and interpretability. It performs well on in-domain pairs like “landscape–awe” but struggles with out-of-domain or unseen cases (e.g., “movement-surprise”).
        \textbf{(b)} Although GRPO improves generalization, its think process is not emotion-oriented and weakly connected to the final answer (e.g., rethinking the last rollout yields “amusement”, while the prediction is “fear”).
    }
	\label{fig:motivation}
    \vspace{-0.3cm}
\end{figure}

\section{Introduction}
\label{sec:intro}


Multimodal Large Language Models (MLLMs)~\cite{LLaVA15_CVPR23,li2024llava} have achieved remarkable progress in visual question answering, visual understanding, and visual generation tasks by leveraging large-scale multimodal data~\cite{chen2024internvl,li2024monkey,wang2024qwen2,rong2025backdoor,SafetySurvey}. However, despite their strong performance on general visual tasks, MLLMs still struggle to capture and interpret emotions effectively~\cite{yang2024emollm,yang2023context}, often generating superficial emotional responses and failing to fully understand complex emotional cues~\cite{cheng2024emotion,EmoVIT,xing2024emo,zhaofacephi,zhang2024microemo,fang2025emoe}.

In the field of visual emotional understanding, many existing studies such as EmoVIT~\cite{EmoVIT}, EmotionL-LaMA~\cite{cheng2024emotion}, AffectGPT~\cite{lian2025affectgpt}, and EmoLLM~\cite{yang2024emollm} primarily adopt Supervised Fine-Tuning (SFT)~\cite{LoRASculpt_CVPR25,huang2025keeping,huang2024learn} to improve model performance on emotional tasks. However, these approaches still have notable limitations in \textbf{generalization and interpretability}~\cite{rajani2025scalpel}. As shown in~\cref{fig:motivation}(a), SFT learns emotional representations by fitting the distribution of the training data, but the limited range of emotional categories and the fixed, predefined label taxonomy constrain the model to discrete emotional types. As a result, the model struggles to capture the continuity, subtle nuances, and contextual variability of visual emotional expressions. This reliance on a closed label space often leads to overfitting and reduces the adaptability of model to unseen visual or affective domains. Moreover, because SFT relies on example-level supervision, its reasoning tends to be pattern-matching rather than genuinely capturing the relationships among emotional factors.
In contrast, applying Reinforcement Learning (RL)~\cite{liu2025reinforcement,zhou2025reinforced} for post-training MLLMs can effectively alleviate these issues. In particular, Group Relative Policy Optimization (GRPO)~\cite{DeepSeekR1_arXiv25,shao2024deepseekmath,GRPO_NeurIPS24,rong2025safegrpo} stands out because, unlike Proximal Policy Optimization (PPO)~\cite{schulman2017proximal} and Direct Preference Optimization (DPO)~\cite{rafailov2023direct,xu2024dpo}, it does not require additional human-annotated reasoning traces for training~\cite{RLwithCoT_arXiv25}. Specifically, GRPO optimizes model behavior based on relative feedback among grouped samples, enabling the model to learn more generalizable emotional reasoning strategies through comparative evaluation. This optimization mechanism allows the model to uncover the latent structures and semantic relationships between visual content and emotional expressions, thereby enhancing its capability in visual emotional understanding and reasoning.

GRPO-based methods typically focus on optimizing the group-relative advantage~\cite{huang2025mapo,yu2025dapo} or improving sampled roll-outs~\cite{TreeRPO_arXiv25,zhang2025r1,SEEDGRPO_arXiv25,yao2025r1} to enhance general capabilities, yet they pay little attention to task-specific adaptation for downstream emotional-understanding tasks. While recent emotion-related reinforcement learning works introduce GRPO into emotional reasoning~\cite{lian2025affectgpt,zhao2025r1}, they largely do so in a superficial manner, reusing its framework without adapting it to the intrinsic nature of emotional cognition. Actually, \ding{182} \textit{\textbf{general GRPO generated reasoning process does not align well with the reasoning patterns required for emotion interpretation}}, especially in visual emotion-understanding scenarios. While the decision-making of GRPO is effective, it fails to reliably capture the intuitive logic underlying human emotional comprehension.

Furthermore, unlike tasks such as mathematical reasoning~\cite{shao2024deepseekmath} or code generation~\cite{robeyns2025improving}, where the relationship between \textit{think} and \textit{answer} is tightly bound, \ding{183} \textit{\textbf{visual emotional understanding tasks lack this direct correspondence between reasoning traces and outputs}}. In mathematical or programming tasks, an incorrect reasoning step almost inevitably leads to an incorrect answer, allowing GRPO to indirectly constrain the reasoning process through answer verification. In contrast, emotional understanding is highly subjective and context-dependent. The reasoning path may diverge from the final \textit{answer} due to individual or contextual variations in emotional interpretation. As shown in \cref{fig:motivation}(b), when we \textit{rethink} the \textit{think}-text from roll-out samples, the inferred emotion often differs from that of the final \textit{answer}, indicating that the correctness of the answer cannot reliably reflect the quality of the reasoning process. Visual emotional tasks require not only perception of visual cues but also comprehension of complex emotional contexts and background knowledge, while maintaining emotional coherence across these cues. Therefore, constraining the answer alone is insufficient to guide the reasoning process effectively, posing a unique challenge for enhancing emotional reasoning in vision-based tasks.

To tackle these challenges, we propose \textbf{R}eflective \textbf{R}einforcement Learning for Emotional \textbf{R}easoning in Multimodal Large Language Models (\ourmethod{}). \textbf{\textit{First}}, we design Structured Emotional Thinking that explicitly guides the model to reason about emotions in a step-by-step manner and constrains its output to follow a specific, interpretable format. This structured formulation helps the model generate coherent emotional reasoning traces rather than fragmented or task-agnostic thoughts. \textbf{\textit{Next}}, we introduce Reflective Emotional Reward, which allows the model to re-evaluate its own reasoning and assess whether its emotional interpretation aligns with visual and contextual cues. By feeding the reasoning back into the model, we apply two rewards: visual-text consistency, ensuring the reasoning is grounded in the visual input, and emotional reasoning validity, enforcing logical soundness and emotional coherence in the inferred emotions. Extensive experiments demonstrate that \ourmethod{} significantly enhances the interpretability and emotional intelligence of multimodal large language models in visual emotional understanding.

The main contributions can be summarized as follows:
\begin{itemize}
\item We propose a Structured Emotional Thinking process that guides MLLMs to perform emotional reasoning in a structured and interpretable manner, improving their ability to understand emotions more human-likely.

\item We introduce a Reflective Emotional Reward mechanism that enables the model to re-evaluate its reasoning and optimize through reflective feedback, ensuring more coherent and grounded emotional reasoning.

\item We conduct extensive experiments demonstrating that \ourmethod{} consistently outperforms previous methods from multiple perspectives.
\end{itemize}
\section{Related Works}
\label{sec:related}

\subsection{Emotion Recognition in MLLMs}
Recent advances in Multimodal Large Language Models (MLLMs)~\cite{LLaVA15_CVPR23,chen2024internvl,li2024monkey,wang2024qwen2,li2024llava} 
have significantly enhanced the joint understanding across visual, textual, and auditory modalities~\cite{liu2025guardreasoner,jin2025two,jin2024learning}, 
and improved the ability to handle a variety of multimodal tasks~\cite{huang2025keeping,huang2024learn,LoRASculpt_CVPR25}. 
Most research in this field focuses on leveraging large-scale pretrained models for general-purpose applications~\cite{liu2024dora,zhao2024galore,han2024onellm,bai2025chat}, including vision-language reasoning~\cite{ScienceQA_NeurIPS22,masry-etal-2022-chartqa,wang2025safety}, image captioning~\cite{COCO_ECCV14,Flickr_TACL14}, 
and visual question answering~\cite{GQA_CVPR19,VQAV2_CVPR17,TextVQA_CVPR19,huang2025be}. 
MLLMs have demonstrated remarkable performance in these tasks~\cite{bi2024visual}, showcasing their ability to integrate and reason across multiple modalities.

However, MLLMs often struggle with emotion-related tasks~\cite{lian2025affectgpt}. These challenges arise from the subjective and context-dependent nature of emotional understanding, 
which requires not only perceptual grounding but also affective reasoning across modalities. To address this issue, several studies have explored supervised fine-tuning MLLMs using emotional datasets~\cite{xing2024emo,yang2024emollm,zhaofacephi,zhang2024microemo}. 
For example, EmoVIT~\cite{EmoVIT} leverages GPT-4 to generate emotion-relevant textual descriptions, helping models better interpret affective cues and capture nuanced emotional expressions. Meanwhile, Emotion-LLaMA~\cite{cheng2024emotion} integrates specialized affective encoders that are designed to capture and interpret emotional signals across multiple modalities, thereby enhancing the capacity of model to understanding emotions from text, audio, and visual inputs. Although these fine-tuning approaches improve performance, they typically require extensive retraining or instruction-based adaptation, leading to high computational costs and limited scalability. Nowadays, increasing attention has been devoted to enhancing the generalization and interpretability of MLLMs~\cite{wei2022chain,lian2023explainable,lian2023explainable2}, which has motivated the exploration of reinforcement learning strategies for downstream emotional understanding. These approaches aim to improve the affective cognition of models and human alignment in open-domain scenarios through more flexible reward signals and adaptive optimization processes.

\subsection{Group Relative Policy Optimization}
\label{sec:relatedworksgrpo}
With the growing adoption of reinforcement learning~\cite{liu2025reinforcement,zhou2025reinforced} in large language models (LLMs)~\cite{touvron2023llama} training~\cite{zhang2024can,gao2024designing}, Group Relative Policy Optimization (GRPO)~\cite{shao2024deepseekmath,GRPO_NeurIPS24} has emerged as a widely used optimization paradigm, originally applied to enhance reasoning and alignment performance in LLMs. Unlike traditional Proximal Policy Optimization (PPO)~\cite{schulman2017proximal}, GRPO optimizes based on in-group relative rewards, generating multiple candidate reasoning trajectories for the same input and computing relative advantages among them. This formulation greatly improves optimization stability and reasoning consistency. Early works such as the DeepSeek-R1~\cite{DeepSeekR1_arXiv25} demonstrated that GRPO can substantially enhance model performance and interpretability in mathematical, logical, and scientific reasoning tasks.

Recently, extensive studies have extended and refined the GRPO framework. For instance, Text-Debiased Hint-GRPO~\cite{huang2025boosting} introduces debiased hint mechanisms to mitigate linguistic bias in multimodal reasoning; R1-VL~\cite{zhang2025r1} adopts a step-wise optimization strategy to stabilize learning across multimodal tasks; and R1-Omni~\cite{zhao2025r1} applies reinforcement learning to omni-modal emotion recognition, verifying its potential in subjective affective reasoning. Moreover, Video-R1~\cite{feng2025video}, VideoChat-R1~\cite{li2025videochat}, and Visual-RFT~\cite{liu2025visual} extend the paradigm to video and vision-centric settings, showcasing its cross-modal scalability.

Nevertheless, in the field of emotion understanding, the application of GRPO remains largely superficial. Most methods merely adapt the general GRPO framework at a surface level~\cite{lian2025affectgpt,zhao2025r1}, without addressing its inherent mismatch with subjective emotional reasoning. Specifically, the reasoning paths generated by general GRPO often diverge from the intuitive logic of human affective reasoning, making it difficult to capture subjective and context-dependent emotional associations. Furthermore, unlike mathematical or coding tasks, where the thinking–answer relationship is tightly coupled, visual emotion understanding lacks such direct correspondence, causing traditional GRPO to struggle with learning stable affective semantic signals. Therefore, developing a GRPO optimization mechanism tailored to emotional understanding is essential for advancing affective reasoning in multimodal large models.
\begin{figure*}[htb]
	\centering{\includegraphics[width=0.97\linewidth]{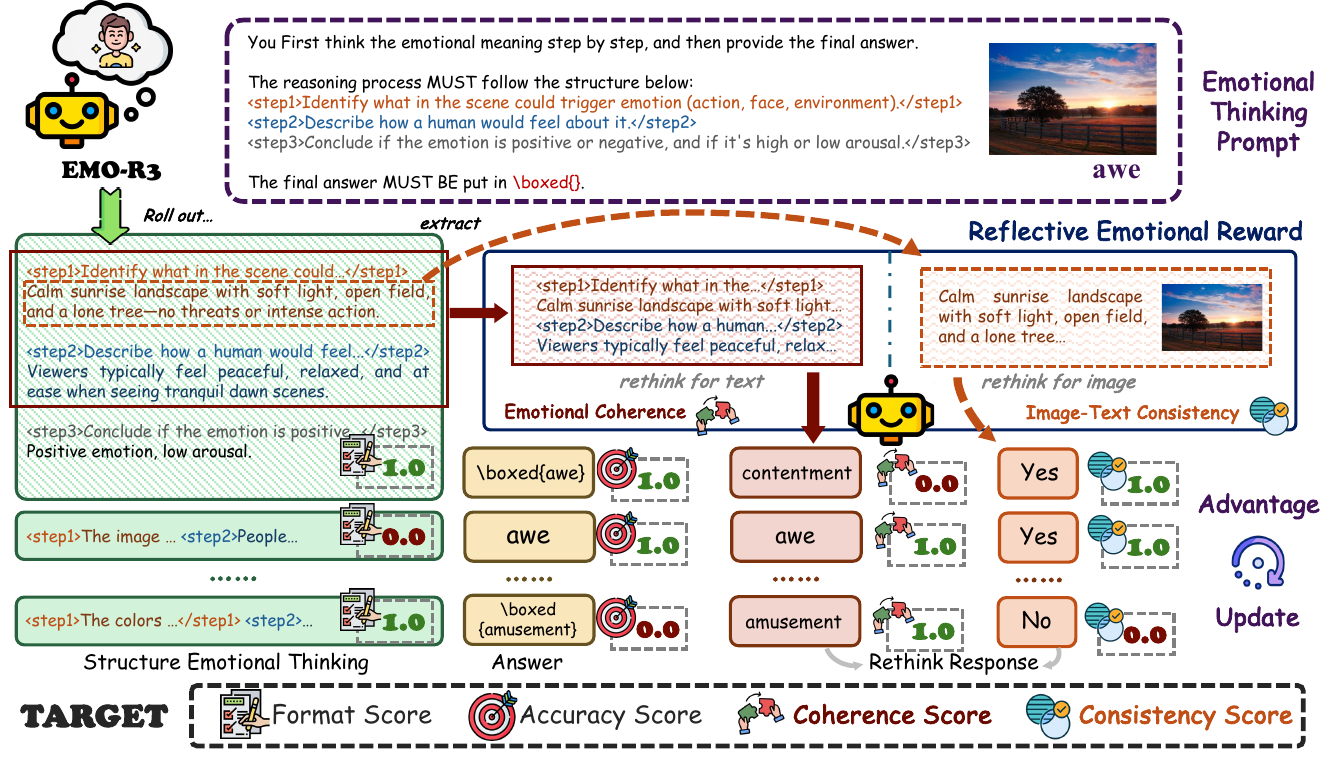}}
    \vspace{-0.1cm}
	\caption{
        \textbf{Architecture illustration} of \ourmethod{}. The upper part presents the Structured Emotional Thinking prompt, which consists of three consecutive thinking steps followed by a final answer. The lower part illustrates the Reflective Emotional Reward mechanism, where multiple rollout samples are evaluated based on image–text consistency and emotional coherence, and are jointly optimized with the original Format and Accuracy rewards under the GRPO framework.
	}
	\label{fig:framework}
	\vspace{-0.4cm}
\end{figure*}

\section{The Proposed Method}
\subsection{Preliminary}
\label{sec:preliminary}
Group Relative Policy Optimization (GRPO) is a variant of Proximal Policy Optimization (PPO). Originally, PPO was designed to enhance mathematical reasoning in large language models. However, GRPO can be effectively adapted to improve visual reasoning and other multimodal capabilities as well. GRPO begins by constructing the current policy model $\pi_{\theta}$ and a reference model $\pi_{\text{old}}$,  
where the latter represents the \textit{old} policy, i.e., the policy from a previous iteration.  
Let $\rho_Q$ denote the distribution of prompts or questions.  
Given a prompt $q \sim \rho_Q$, the model samples a group of outputs $o_1, o_2, \ldots, o_G$ from the old policy $\pi_{\text{old}}$.  
The policy $\pi_{\theta}$ is then optimized by maximizing the following objective function:
\begin{align}
\mathcal{J}_{\mathrm{GRPO}}(\theta) 
&= \mathbb{E}_{q \sim \rho_Q} \mathbb{E}_{o \sim \pi_{\text{old}}(\cdot|q)} 
\Bigg[
\frac{1}{G} \sum_{i=1}^{G} 
f_{\epsilon}\!\left(
\frac{\pi_{\theta}(o_i|q)}{\pi_{\text{old}}(o_i|q)}, \hat{A}_i
\right)
\Bigg] \nonumber \\
&\quad - \beta\, \mathbb{D}_{KL}\!\left[\pi_{\theta} \| \pi_{\text{ref}}\right],
\end{align}

where $\beta$ is the hyperparameter, and
$f_\epsilon(x, y) = \min(xy, \text{clip}(x, 1 - \epsilon, 1 + \epsilon)y)$.
$\hat{A}i$ denotes the advantage, which is calculated based on the relative rewards of the outputs within each group.
More specifically, for each question $q$, a group of outputs ${o_1, o_2, \ldots, o_G}$ is sampled from the old policy model $\pi_{\text{old}}$.
A reward function $R$ is then used to score these outputs, yielding $G$ rewards $\mathbf{r} = {r_1, r_2, \ldots, r_G}$, where $r_i = \mathcal{R}(q, o_i)$.
The mean reward is computed as $\mu = \frac{1}{G} \sum_{i=1}^{G} r_i$, and the standard deviation is defined as
$\sigma = \sqrt{\frac{1}{G} \sum_{i=1}^{G} (r_i - \mu)^2}$.
The normalized advantage for the $i^{\text{th}}$ rollout is then defined as
$\hat{A}_i=\frac{r_i-\mu}{\sigma}$.
This normalization ensures that the advantage values have zero mean and unit variance within each group, stabilizing gradients and promoting consistent optimization dynamics.

\subsection{Structured Emotional Thinking (SET)}

\paragraph{Motivation.}
Although GRPO has shown effectiveness in improving general reasoning abilities of large multimodal models, its prompting design for the thinking stage is often minimal, typically consisting of a single instruction such as \textit{think}.
This one-step thinking cue is task-agnostic and provides no explicit guidance on how emotional reasoning should be organized. In emotional understanding tasks, such a simplistic prompt frequently causes the model to generate fragmented or inconsistent emotional reasoning traces that fail to capture the subtle relationships between visual cues and human affective appraisal.
In visual emotional understanding, where the mapping between perception and emotion is complex and context dependent, a single \textit{think} instruction is insufficient to elicit coherent or human-aligned reasoning.

\paragraph{Designed Prompt.}
To achieve interpretable and human-like emotional understanding, we propose Structured Emotional Thinking (SET), a module that guides the model to perform emotion reasoning in a structured, step-by-step manner before generating the final prediction.

Concretely, SET constrains the reasoning process into three explicit stages, mirroring how humans interpret emotions in visual scenes:
\definecolor{keywordcolor}{RGB}{178,34,34}
\definecolor{c2framework}{RGB}{250, 249, 222} 
\begin{mdframed}[backgroundcolor=c2framework!40, linewidth=2pt, linecolor=c2framework!250, roundcorner=20pt]
\textbf{\textcolor{keywordcolor}{Structured Emotional Thinking:}}
\begin{itemize}
    \item \textbf{Emotional Trigger Identification:} Detect which elements in the scene (objects, actions, environments, or facial cues) may trigger emotional responses.
    \item \textbf{Human Emotional Reflection:} Describe how a human observer would emotionally respond to these elements.
    \item \textbf{Emotional Conclusion:} Determine whether the overall emotion is positive or negative, and assess its arousal level (e.g., calm vs. excited).
\end{itemize}
\end{mdframed}

Given a multimodal input pair $(I, T)$, the model generates a structured reasoning output $o = \{s_1, s_2, s_3, \hat{\mathcal{E}}\}$ corresponding to the three stages and the final \textit{answer} of emotional reasoning:
\begin{equation}
    o = \mathcal{M}_{\theta}(I, T), \quad \hat{\mathcal{E}} = \mathcal{F}_a(o),
\end{equation}
where $\mathcal{M}_{\theta}$ denotes the multimodal reasoning model parameterized by $\theta$, 
and $\mathcal{F}_a(\cdot)$ outputs the final \textit{answer} enclosed in \texttt{\textbackslash boxed\{\}}.

\paragraph{General Reward.}
Following the GRPO setting, we define two reward terms to guide the optimization of the structured emotional reasoning model.

The format reward $\mathcal{R}_{\text{format}}$ measures whether the generated reasoning sequence adheres to the prescribed structure. 
Specifically, it checks whether each reasoning step $s_i$ corresponds to the expected stage and whether the final \textit{answer} is correctly enclosed in \texttt{\textbackslash boxed\{\}}:
\[
    \mathcal{R}_{\text{format}} = 
    \begin{cases}
        1, & \text{if the \texttt{step} and \texttt{box} format are correct;} \\
        0, & \text{otherwise.}
    \end{cases}
\]
Meanwhile, the accuracy reward $\mathcal{R}_{\text{acc}}$ evaluates whether the predicted emotional label $\hat{\mathcal{E}}$ aligns with the ground-truth emotion label $\mathcal{E}^*$:
\[
    \mathcal{R}_{\text{acc}} = 
    \begin{cases}
        1, & \text{if } \hat{\mathcal{E}} = \mathcal{E}^*; \\
        0, & \text{otherwise.}
    \end{cases}
\]

These two general rewards serve as the foundational supervision signal that initiates the optimization of the structured emotional thinking model under the GRPO framework, ensuring that the model first learns to produce structurally valid and semantically accurate emotional reasoning before incorporating higher-level reflective objectives.

\begin{algorithm}[t]
\small  
\caption{\ourmethod{}}
\label{alg:set-rer}
\SetAlgoLined
\SetNoFillComment
\DontPrintSemicolon
\SetKwInput{KwIn}{Input}
\SetKwInput{KwOut}{Output}

\KwIn{Dataset $\mathcal{D} = \{(I, T, \mathcal{E}^*)\}$, pretrained model $\mathcal{M}_{\theta}$, rollout number $G$, coefficients $\lambda_1, \lambda_2$}
\KwOut{Optimized model $\mathcal{M}'_{\theta}$}

\ForEach{$(I, T, \mathcal{E}^*) \in \mathcal{D}$}{
    {\footnotesize \color{DarkBlue}{\tcc{Generate multiple reasoning outputs with structured prompt}}}
    $\{o_1, o_2, \ldots, o_G\} \sim \pi_{\text{old}}(\cdot | I, T)$ \;
    
    {\footnotesize \tcc{Compute rewards for each rollout}}
    \For{$i = 1$ \KwTo $G$}{
        {\footnotesize \color{DarkBlue}{\tcc{(1) General Reward}}}
        Parse $o_i = \{s_1, s_2, s_3, \hat{\mathcal{E}}\}$\;
        $\mathcal{R}^{(i)}_{\text{format}} = \mathbb{I}(\text{step and box format correct})$\;
        $\mathcal{R}^{(i)}_{\text{acc}} = \mathbb{I}(\hat{\mathcal{E}} = \mathcal{E}^*)$\;
        
        {\footnotesize \color{DarkBlue}{\tcc{(2) Reflective Emotional Reward}}}
        Image-text consistency:\\
        $\hat{y}^{(i)}_{\text{cons}} = \mathcal{M}(I, s_1, \mathcal{P}_{\text{cons}})$, 
        $\mathcal{R}^{(i)}_{\text{cons}} = \mathbb{I}(\hat{y}^{(i)}_{\text{cons}} = \text{Yes})$\;
        
        Emotional coherence:\\
        $\hat{y}^{(i)}_{\text{coh}} = \mathcal{M}(s_{1,2}, \mathcal{P}_{\text{coh}})$,
        $\mathcal{R}^{(i)}_{\text{coh}} = \mathbb{I}(\hat{y}^{(i)}_{\text{coh}} = \mathcal{E}^*)$\;
        
        $\mathcal{R}^{(i)}_{\text{RER}} = \tfrac{1}{2}(\mathcal{R}^{(i)}_{\text{cons}} + \mathcal{R}^{(i)}_{\text{coh}})$\;
        
        {\footnotesize \color{DarkBlue}{\tcc{(3) Combine into overall reward}}}
        $\mathcal{R}^{(i)}_{\text{overall}} =
        (1 - \lambda_1 - \lambda_2)\mathcal{R}^{(i)}_{\text{acc}}
        + \lambda_1\mathcal{R}^{(i)}_{\text{RER}}
        + \lambda_2\mathcal{R}^{(i)}_{\text{format}}$\;
    }
    
    {\footnotesize \color{DarkBlue}{\tcc{Compute advantage and update model via GRPO}}}
    Normalize rewards within group to obtain $\hat{A}_i$;\\
    Update policy parameters $\theta$ using GRPO objective with advantages $\hat{A}_i$.
}

\Return{$\mathcal{M}'_{\theta}$}
\end{algorithm}

\subsection{Reflective Emotional Reward (RER)}

\paragraph{Motivation.}
Although the Designed Prompt provides a structured framework for emotional reasoning, it cannot ensure that the generated reasoning is visually consistent with the textual content or emotionally coherent. Meanwhile, the general GRPO formulation lacks effective constraints on the \textit{think} process; by supervising only the final \textit{answer}, it fails to effectively select high-quality reasoning samples.

\paragraph{Image-Text Consistency Reward.}
The image-text consistency reward enforces alignment between the generated reasoning and the visual content of the image.

In this process, we extract only \texttt{step1} from the model output $o$, denoted as $s_1 = \mathcal{F}_1(o)$, and feed it back into the model together with the image $I$. The prompt for this reflective process is denoted as $\mathcal{P}_\text{cons}$:
\begin{mdframed}[backgroundcolor=c2framework!40, linewidth=2pt, linecolor=c2framework!250, roundcorner=20pt]
\centering
\textbf{\textit{Can the following text describe the image?}}
\end{mdframed}

The model then produces a reflective output:
\begin{equation}
    \hat{y}_{\text{cons}} = \mathcal{M}(I, s_1, \mathcal{P}_\text{cons}),
\end{equation}
where the response can be either ``Yes'' or ``No''. The corresponding reward is defined as:
\[
\mathcal{R}_{\text{cons}} = 
\begin{cases} 
1, & \text{if } \hat{y}_{\text{cons}} = \text{Yes;} \\
0, & \text{if } \hat{y}_{\text{cons}} = \text{No.}
\end{cases}
\]

This reward encourages the model to generate reasoning that is both emotionally coherent and visually grounded, ensuring stronger alignment between textual descriptions and image semantics.

\paragraph{Emotional Coherence Reward.}
The emotional coherence reward aims to evaluate whether the reasoning process maintains consistency with the ground-truth emotion label.

In this process, we extract \texttt{step1} and \texttt{step2} from the model-generated reasoning, denoted as $s_{1,2} = \mathcal{F}_{1,2}(o)$, which is fed back into the model for reflection. The prompt for this reflective process is denoted as $\mathcal{P}_\text{coh}$:
\begin{mdframed}[backgroundcolor=c2framework!40, linewidth=2pt, linecolor=c2framework!250, roundcorner=20pt]
\centering
\textbf{\textit{Which emotion best describes the text above?}}
\end{mdframed}

The model then produces a reflective output:
\begin{equation}
    \hat{y}_{\text{coh}} = \mathcal{M}(R_{\text{input}}, \mathcal{P}_\text{coh}),
\end{equation}
where $\hat{y}_{\text{coh}}$ represents the emotion label predicted by the model during the reflective stage. 
We then compare $\hat{y}_{\text{coh}}$ with the ground-truth emotion label $\mathcal{E}^*$. 
The emotional coherence reward is defined as:
\[
\mathcal{R}_{\text{coh}} = 
\begin{cases} 
1, & \text{if } \hat{y}_{\text{coh}} = \mathcal{E}^*; \\
0, & \text{otherwise.}
\end{cases}
\]

This reward encourages the model to generate reasoning that is emotionally consistent with the ground-truth label, thereby improving the emotional coherence and interpretability of the reasoning process.

\subsection{Overall Reward and Discussion}
\paragraph{Overall Reward.}
The final optimization objective integrates all the previously defined reward components into a unified formulation.  
Specifically, the reflective emotional reward is obtained by averaging the image-text consistency reward and the emotional coherence reward:
\begin{equation}
    \mathcal{R}_{\text{RER}} = \frac{\mathcal{R}_{\text{cons}} + \mathcal{R}_{\text{coh}}}{2}.
\end{equation}

Subsequently, the overall reward used for GRPO optimization is defined as a weighted combination of the accuracy reward, the reflective emotional reward, and the format reward, which is calculated as follows:
\begin{equation}
    \mathcal{R}_{\text{overall}} = 
    (1 - \lambda_1 - \lambda_2)\, \mathcal{R}_{\text{acc}}
    + \lambda_1\, \mathcal{R}_{\text{RER}}
    + \lambda_2\, \mathcal{R}_{\text{format}},
\end{equation}
where $\lambda_1$ and $\lambda_2$ are balancing coefficients that control the relative contributions of emotional coherence and structural correctness.

By combining these complementary rewards, the training process promotes reasoning traces that are not only visually grounded and emotionally coherent but also maintain interpretable, human-aligned structure.

\paragraph{Discussion on Cold-Start-Emo.}
In our framework, a key question is whether Supervised Fine-Tuning (SFT) should be introduced as a cold-start stage before GRPO optimization. Unlike factual reasoning or visual question answering tasks, emotion recognition inherently involves subjectivity.
Pretrained MLLMs often carry emotional priors derived from large-scale corpora, which reflect general or culture-dependent affective tendencies. These priors may deviate considerably from the labeling schemes of specific downstream datasets. If GRPO training is conducted without any prior alignment, such a mismatch can cause the model to repeatedly generate reasoning traces that are inconsistent with the dataset annotations, leading to sparse reward signals and consequently weakening optimization stability.

Inspired by previous studies, many works perform cold start with Chain-of-Thought (CoT)-annotated datasets prior to GRPO. The main motivation behind this design is to endow the model with an initial \textit{thinking} ability, enabling more effective reasoning optimization later. Our motivation, however, is different: instead of enhancing the reasoning-chain capability of model, we aim to alleviate the training difficulty caused by subjective bias in emotional understanding tasks.

To this end, we explore a lightweight SFT-based Cold Start for Emotional Reasoning (Cold-Start-Emo) using a small number of samples without CoT annotations. This stage requires no additional reasoning chains; rather, a small set of task-specific examples is used to help the model preliminarily learn the task format, emotional label system, and expression patterns. Such initialization enables an early-stage alignment between the pretrained priors and the target task distribution. Empirical results demonstrate that this initialization allows the model to generate higher-quality rollouts more stably during subsequent GRPO training, mitigating reward sparsity and ultimately improving both the coherence and accuracy of emotional reasoning.

\definecolor{mygray}{gray}{.9}
\definecolor{myyellow}{RGB}{252, 240, 199}
\begin{table*}[t]
	\centering
 	\caption{\textbf{Performance comparison with the state-of-the-art GRPO variants} on the emotional reasoning tasks across in-domain and out-of-domain settings. * denotes models without post-training. Datasets marked with the superscript $I$, \eg EmoSet$^I$ and Emotion6$^I$, denote the in-domain training dataset. We mark the Best in bold across different methods. Please refer to \cref{subsec:comp_exp} for details.}
     \vspace{-0.15cm}
	\setlength{\tabcolsep}{6.5pt}
    \renewcommand{\arraystretch}{1.1}
	\scalebox{0.95}{
	\begin{tabular}{c||c|c|cc||c|cc||ccc}
		\hline\thickhline
        \rowcolor{mygray}
		Methods & Roll-out & EmoSet$^I$ & Emotion6 & WebEmo & Emotion6$^I$ & EmoSet & WebEmo & $\mathcal{A}^I$ & $\mathcal{A}^O$ & $\mathcal{A}$ \\
		\hline
		\hline
        \multicolumn{10}{l}{\textcolor{gray!80}{\textit{LLaVA1.5-7B}}}\\ 
        \rowcolor{gray!8}
        Vanilla* & - & 52.77 & 48.32 & 25.56 & 48.32 & 52.77 & 25.56 & 50.55 & 38.05 & 42.22 \\
        SEPM* & - & 56.04 & 54.21 & 42.39 & 54.21 & 56.04 & 42.39 & 55.13 & 48.76 & 50.88 \\
        \hline
        \multicolumn{10}{l}{\textcolor{gray!80}{\textit{Qwen2.5-VL-3B-Instruct}}}\\ 
        \rowcolor{gray!8}
        Vanilla* & - & 51.55 & 50.00 & 40.65 & 50.00 & 51.55 & 40.65 & 50.77 & 45.71 & 47.40 \\
        SFT & - & 77.15 & 34.51 & 17.75 & 69.53 & 26.45 & 37.65 & 73.34 & 29.09 & 43.84 \\
        \hline
        \rowcolor{gray!8}
        GRPO & & 74.60 & 60.10 & 49.50 & 70.88 & 59.90 & 44.85 & 72.74 & 53.59 & 59.97 \\
        DAPO & & 68.99 & 56.90 & 49.80 & 68.56 & 59.95 & 45.50 & 68.78 & 53.04 & 58.28 \\
        \rowcolor{myyellow}
        \ourmethod{} & \multirow{-3}{*}{\textit{4}} & 75.50 & 60.44 & 50.45 & 70.71 & 60.70 & 45.20 & \textbf{73.10} & \textbf{54.20} & \textbf{60.50} \\
        \hline
        GRPO &  & 75.45 & 57.91 & 49.40 & 69.87 & 60.30 & 42.05 & 72.66 & 52.42 & 59.16 \\
        \rowcolor{gray!8}
        DAPO &  & 70.21 & 55.72 & 48.80 & 62.39 & 58.05 & 46.30 & 66.30 & 52.22 & 56.91 \\
        \rowcolor{myyellow}
        \ourmethod{} & \multirow{-3}{*}{\textit{8}} & 76.40 & 59.26 & 49.70 & 71.72 & 61.80 & 43.65 & \textbf{74.06} & \textbf{53.60} & \textbf{60.42} \\

	\end{tabular}}

    \label{tab:comp_sota}
    \vspace{-0.15cm}
\end{table*}

\section{Experiments}
\subsection{Experimental Setup}
\paragraph{Environment and Dataset.}
Our training framework uses EasyR1, and the testing framework uses NoisyRollout.
Our experiments use three emotion datasets: EmoSet~\cite{yang2023emoset} (8 categories), Emotion6~\cite{peng2015mixed} (6 categories), and WebEmo~\cite{panda2018contemplating} (7 categories). We train separately on the EmoSet and Emotion6 datasets, while other datasets are used as out-of-domain tests. To enhance efficiency, we randomly cropped the training and testing sets of each dataset (2,000 samples).

\paragraph{Architecture and Counterparts.}
We utilize the popular open-source Qwen2.5-VL-3B-Instruct~\cite{bai2025qwen2} as the base (Vanilla) model, which exhibits strong foundational capabilities well-suited for subsequent RL training. We further compare our approach with the training-free method SEPM~\cite{fang2025sepm}, as well as GRPO~\cite{shao2024deepseekmath} and DAPO~\cite{yu2025dapo}, to validate its effectiveness.

\paragraph{Implement Details.}
The experimental results are obtained at the same step (when convergence is reached), with the hyperparameters $\lambda_1$ and $\lambda_2$ both set to 0.1. The learning rate for the experiment is set to 2.0e-6. To eliminate randomness, we ran the experiment three times and reported the median result. The experiments are conducted on a total of 8 NVIDIA H20 GPUs, each with 96GB of memory.

\paragraph{Evaluation Metrics. }
We evaluate both in-domain and out-of-domain performance. For each dataset, we use Accuracy (ACC) as the evaluation metric. We further compute the average performance for in-domain ($\mathcal{A}^I$) and out-of-domain ($\mathcal{A}^O$) evaluations. Finally, we take the average of all these results to obtain the overall performance ($\mathcal{A}$).

\begin{figure}[t]
    \centering
	\includegraphics[width=1.02\linewidth]{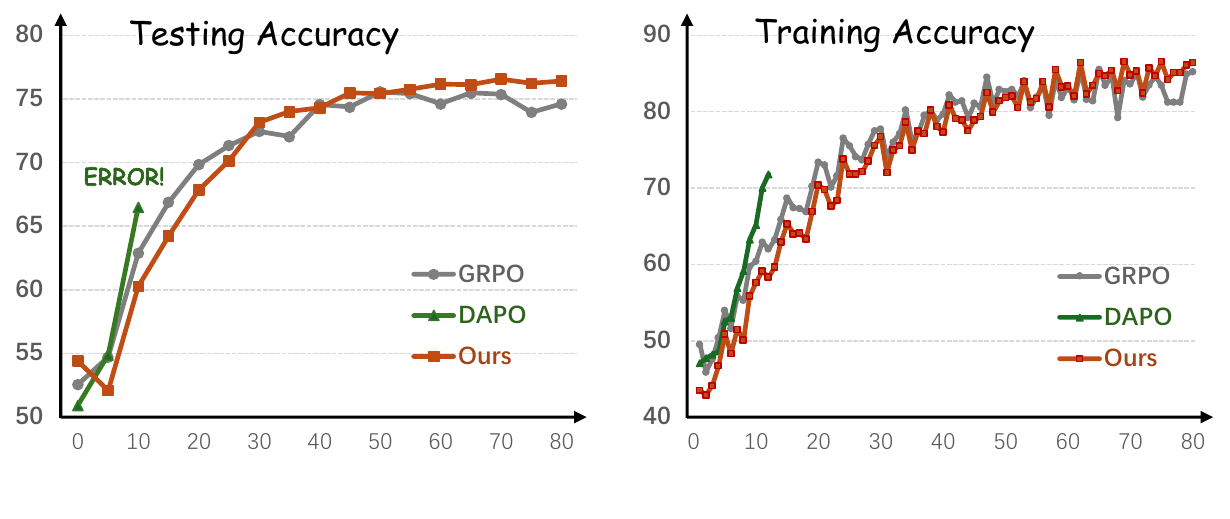}
    \vspace{-0.8cm}
    \caption{
    	Training and testing accuracy during the training process. DAPO fails to conduct complete training. A more detailed analysis of this failure is provided in \cref{subsec:comp_exp}.
    }
	\label{fig:acc_vis}
    \vspace{-0.3cm}
\end{figure}

\subsection{Comparison Experiments}
\label{subsec:comp_exp}

\noindent\textbf{Comparison with State-of-the-art.} We compare the proposed approach with GRPO variants on both in-domain and out-of-domain emotional datasets, as well as with several training-free methods. As reported in \cref{tab:comp_sota}, \ourmethod{} consistently achieves the highest overall accuracy across both 4-rollout and rollout-8 settings, demonstrating its ability to enhance emotional reasoning. Compared to these baselines, our method yields higher in-domain performance, reflecting better alignment with emotional cues learned from the training distributions. Meanwhile, the gain in out-of-domain accuracy shows that our learning strategy mitigates overfitting and improves robustness to domain shift. Notably, DAPO fails to conduct complete training, as shown in \cref{fig:acc_vis}. The failure stems from a fundamental mismatch between the filtering strategy of DAPO and the discrete nature of emotional reasoning evaluation, where the binary reward structure conflicts with the continuous filtering criteria, leading to sample depletion and training instability.

\noindent\textbf{Experiment on Cold-Start-Emo.} We explore the Cold-Start-Emo under the rollout-8 setting. The Cold-Start-Emo is designed to provide early alignment and stabilize the learning process for emotional reasoning. As shown in \cref{tab:coldstartemo}, the integration of Cold-Start-Emo significantly outperforms \ourmethod{} and all other baselines on the in-domain dataset, and it achieves the highest overall average accuracy on the out-of-domain datasets. This empirical evidence validates that Cold-Start-Emo is a highly effective initialization strategy that generates higher-quality rollouts and mitigating reward sparsity during subsequent GRPO training.

\begin{table}[t]
	\centering
 	\caption{\textbf{Experiment on Cold-Start-Emo.} \ourmethod{}$^{\#}$ denotes \ourmethod{} with additional Cold-Start-Emo module. See \cref{subsec:comp_exp}.}
    \vspace{-0.15cm}
    \label{tab:coldstartemo}
	\setlength{\tabcolsep}{7.4pt}
    \renewcommand{\arraystretch}{1.1}
	\scalebox{0.95}{
	\begin{tabular}{c|c|cc|c}
		\hline\thickhline
        \rowcolor{mygray}
		Methods & EmoSet$^I$ & Emotion6 & WebEmo & \ \ \ \ \ $\mathcal{A}$ \ \ \ \ \  \\
		\hline
        SFT & 77.15 & 34.51 & 17.75 & 43.14 \\
        \rowcolor{gray!8}
        GRPO & 75.45 & 57.91 & 49.40 & 60.92 \\
        \ourmethod{} & 76.40 & \textbf{59.26} & 49.70 & 61.79 \\
        \rowcolor{gray!8}
        \ourmethod{}$^{\#}$ & \textbf{77.81} & 58.59 & \textbf{50.00} & \textbf{62.13} \\    
	\end{tabular}}
    \vspace{-0.1cm}
\end{table}

\begin{table}[t]
	\centering
 	\caption{\textbf{Ablative study} of Structured Emotional Thinking (SET) and Reflective Emotional Reward (RER). Please see \cref{subsec:abla}.}
    \vspace{-0.2cm}
	\setlength{\tabcolsep}{6.8pt}
    \renewcommand{\arraystretch}{1.1}
	\scalebox{0.95}{
	\begin{tabular}{cc|c|cc|c}
		\hline\thickhline
        \rowcolor{mygray}
		  SET & RER & EmoSet$^I$ & Emotion6 & WebEmo & \ \ \ \ $\mathcal{A}$ \ \ \ \ \\
		\hline
        $\circ$ & $\circ$ & 75.45 & 57.91 & 49.40 & 60.92 \\
        \rowcolor{gray!8}
        $\bullet$ & $\circ$ & 76.02 & 58.42 & 49.20 & 61.21 \\
        $\bullet$ & $\bullet$ & 76.40 & 59.26 & 49.70 & 61.79 \\
	\end{tabular}}
    \label{tab:abla}
    \vspace{-0.3cm}
\end{table}

\begin{figure*}[htb]
	\centering{\includegraphics[width=0.97\linewidth]{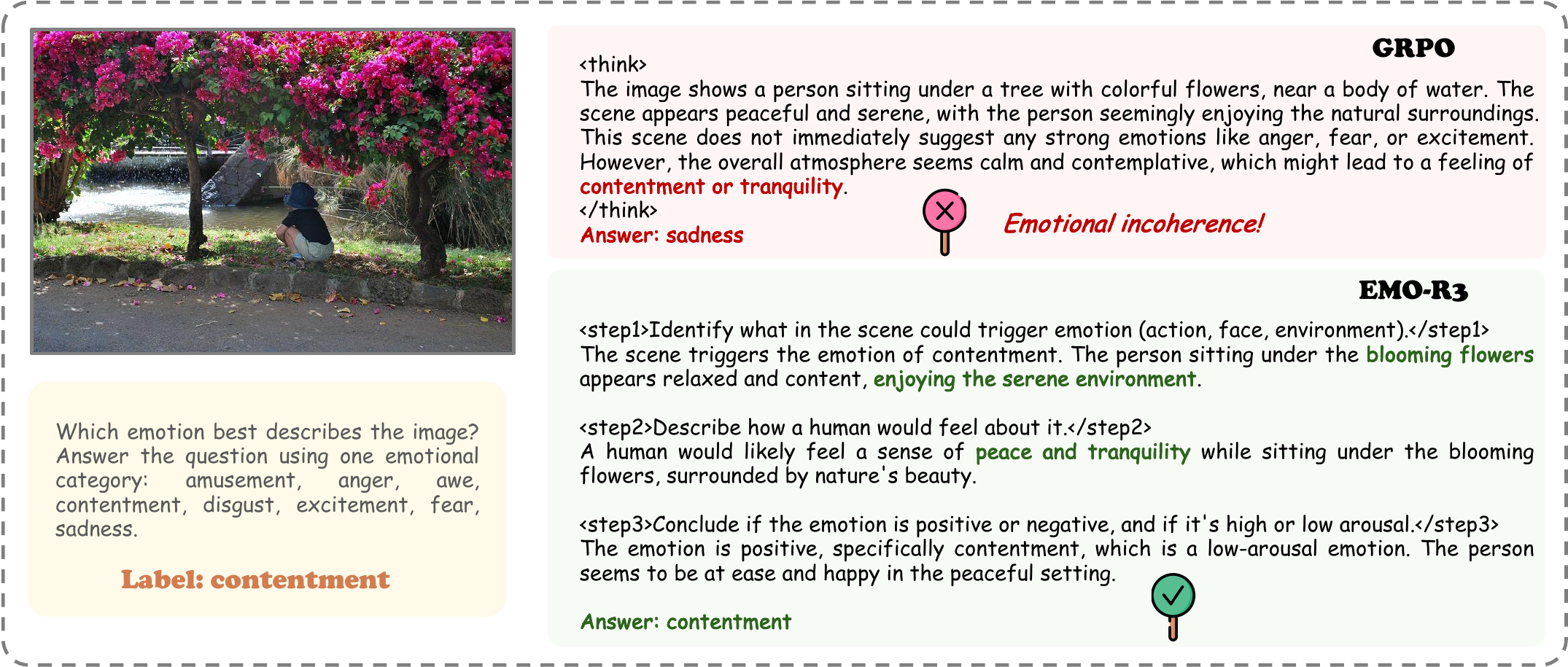}}
    \vspace{-0.2cm}
	\caption{
        \textbf{Case study} between GRPO and \ourmethod{} on the EmoSet dataset. Please see \cref{subsec:casestudy} for details.
	}
	\label{fig:case}
    \vspace{-0.2cm}
\end{figure*}

\begin{figure}[t]
	\centering{\includegraphics[width=1.02\linewidth]{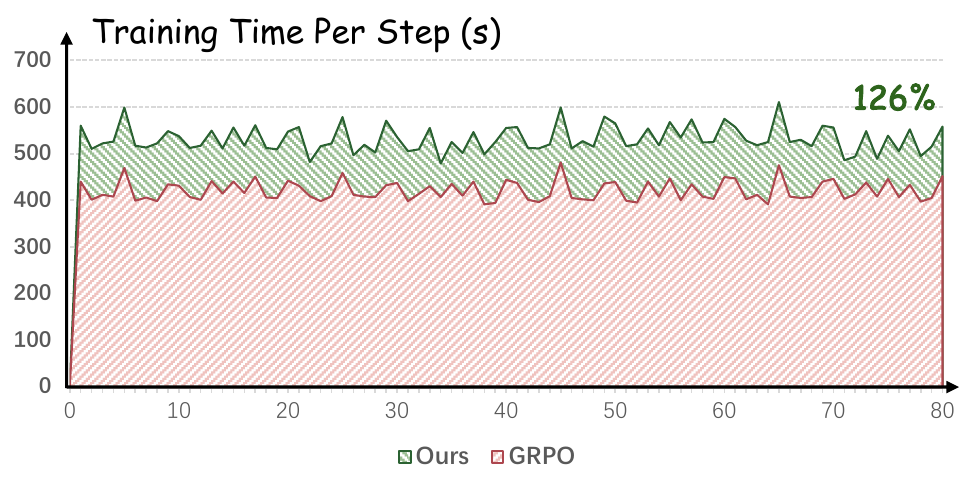}}
    \vspace{-0.7cm}
	\caption{
        \textbf{Efficiency analysis} on the training process. See \cref{subsec:time}.
	}
	\label{fig:time}
	\vspace{-0.3cm}
\end{figure}

\subsection{Ablation Experiments}
\label{subsec:abla}
In \cref{tab:abla}, we begin by validating the effectiveness of our proposed components through their incremental integration. As shown, incorporating Structured Emotional Thinking (SET) consistently enhances performance compared with the baseline, indicating that explicitly organizing the emotional reasoning procedure helps the model produce more coherent, fine-grained, and interpretable emotion representations. When the Reflective Emotional Reward (RER) is further introduced, the model achieves additional improvements, suggesting that reflective self-assessment encourages the model to better align its emotional reasoning with the underlying multimodal evidence. Taken together, these findings demonstrate that the combined use of SET and RER not only improves the interpretability of the reasoning process but also substantially enhances the emotional intelligence of MLLMs.

\subsection{Case Study}
\label{subsec:casestudy}
We evaluated the methods on the EmoSet dataset and selected a representative case for detailed analysis. We observed that the naive GRPO failed to attend to the most emotionally salient regions (\textit{blooming flowers}), and its think and answer components exhibited emotional incoherence. In contrast, our proposed method (\ourmethod{}) effectively addresses this issue by producing emotionally coherent reasoning and predictions. This case demonstrates that \ourmethod{} can accurately capture subtle affective cues and exhibit emotionally coherent reasoning, thereby leading to better emotional understanding and overall performance.

\subsection{Efficiency Analysis}
\label{subsec:time}
Considering that we introduce an additional reflection stage, we conducts an efficiency analysis on the training process under the rollout-8 setting on the EmoSet dataset. As shown in~\cref{fig:time}, although our method introduces a certain amount of extra computation time, it does not lead to a proportional increase in training cost. Moreover, our inference process does not require the reflection module, so it introduces \textbf{no additional inference-time cost}. Thus, our approach maintains high efficiency while achieving better performance.
\section{Conclusion}
In this work, we investigate the challenges of interpretability and generalization faced by Multimodal Large Language Models (MLLMs) in emotional understanding. Although general GRPO-based approaches can partially alleviate these issues, existing models still struggle to accurately capture the subtle, subjective, and context-dependent nature of human emotions. To address this gap, We propose Reflective Reinforcement Learning for Emotional Reasoning in Multimodal Large Language Models (\ourmethod{}). Our method integrates a Structured Emotional Thinking module to guide step-by-step affective reasoning and employs a Reflective Emotional Reward mechanism to ensure visual–textual consistency and coherent emotional expression. Without requiring additional annotations, \ourmethod{} significantly improves both the interpretability and generalization of MLLMs. We believe this work offers new insights for developing emotionally intelligent and human-aligned MLLMs. Building upon this foundation, future research may further explore the generalization of emotion recognition with reasoning in more complex multimodal scenarios, including sequential or interactive task settings. 

{
{
    \small
    \bibliographystyle{ieeenat_fullname}
    \bibliography{main}

@String(ACL  = {ACL})

@String(ACMMMW = {ACM MM Workshop})

@String(CVPR  = {CVPR})

@String(ECCV  = {ECCV})

@String(ICCV  = {ICCV})

@String(ICLRW = {ICLR Workshop})

@String(ICML  = {ICML})

@String(NeurIPS  = {NeurIPS})

@string(TACL = {TACL})

@inproceedings{yang2023context,
  title={Context de-confounded emotion recognition},
  author={Yang, Dingkang and Chen, Zhaoyu and Wang, Yuzheng and Wang, Shunli and Li, Mingcheng and Liu, Siao and Zhao, Xiao and Huang, Shuai and Dong, Zhiyan and Zhai, Peng and others},
  booktitle=CVPR,
  pages={19005--19015},
  year={2023}
}

@inproceedings{LLaVA15_CVPR23,
    author={Liu, Haotian and Li, Chunyuan and Li, Yuheng and Lee, Yong Jae},
    title={Improved Baselines with Visual Instruction Tuning}, 
    booktitle   = CVPR,
    year        = {2023}
}

@inproceedings{chen2024internvl,
  title={Internvl: Scaling up vision foundation models and aligning for generic visual-linguistic tasks},
  author={Chen, Zhe and Wu, Jiannan and Wang, Wenhai and Su, Weijie and Chen, Guo and Xing, Sen and Zhong, Muyan and Zhang, Qinglong and Zhu, Xizhou and Lu, Lewei and others},
  booktitle=CVPR,
  pages={24185--24198},
  year={2024}
}

@inproceedings{li2024monkey,
  title={Monkey: Image resolution and text label are important things for large multi-modal models},
  author={Li, Zhang and Yang, Biao and Liu, Qiang and Ma, Zhiyin and Zhang, Shuo and Yang, Jingxu and Sun, Yabo and Liu, Yuliang and Bai, Xiang},
  booktitle=CVPR,
  pages={26763--26773},
  year={2024}
}

@article{wang2024qwen2,
  title={Qwen2-vl: Enhancing vision-language model's perception of the world at any resolution},
  author={Wang, Peng and Bai, Shuai and Tan, Sinan and Wang, Shijie and Fan, Zhihao and Bai, Jinze and Chen, Keqin and Liu, Xuejing and Wang, Jialin and Ge, Wenbin and others},
  journal={arXiv preprint arXiv:2409.12191},
  year={2024}
}

@article{li2024llava,
  title={Llava-onevision: Easy visual task transfer},
  author={Li, Bo and Zhang, Yuanhan and Guo, Dong and Zhang, Renrui and Li, Feng and Zhang, Hao and Zhang, Kaichen and Li, Yanwei and Liu, Ziwei and Li, Chunyuan},
  journal={arXiv preprint arXiv:2408.03326},
  year={2024}
}

@article{bai2025qwen2,
  title={Qwen2. 5-vl technical report},
  author={Bai, Shuai and Chen, Keqin and Liu, Xuejing and Wang, Jialin and Ge, Wenbin and Song, Sibo and Dang, Kai and Wang, Peng and Wang, Shijie and Tang, Jun and others},
  journal={arXiv preprint arXiv:2502.13923},
  year={2025}
}

@inproceedings{zhang2024microemo,
  title={MicroEmo: Time-Sensitive Multimodal Emotion Recognition with Subtle Clue Dynamics in Video Dialogues},
  author={Zhang, Liyun and Luo, Zhaojie and Wu, Shuqiong and Nakashima, Yuta},
  booktitle=ACMMMW,
  pages={110--115},
  year={2024}
}

@inproceedings{han2024onellm,
  title={Onellm: One framework to align all modalities with language},
  author={Han, Jiaming and Gong, Kaixiong and Zhang, Yiyuan and Wang, Jiaqi and Zhang, Kaipeng and Lin, Dahua and Qiao, Yu and Gao, Peng and Yue, Xiangyu},
  booktitle=CVPR,
  pages={26584--26595},
  year={2024}
}

@inproceedings{COCO_ECCV14,
    title={Microsoft COCO: Common Objects in Context},
    author={Lin, Tsung-Yi and Maire, Michael and Belongie, Serge and Hays, James and Perona, Pietro and Ramanan, Deva and Doll{\'a}r, Piotr and Zitnick, C Lawrence},
    booktitle=ECCV,
    pages={740--755},
    year={2014}
}

@article{Flickr_TACL14,
  title={From image descriptions to visual denotations: New similarity metrics for semantic inference over event descriptions},
  author={Young, Peter and Lai, Alice and Hodosh, Micah and Hockenmaier, Julia},
  journal=TACL,
  volume={2},
  pages={67--78},
  year={2014}
}

@inproceedings{ScienceQA_NeurIPS22,
  title = {Learn to Explain: Multimodal Reasoning via Thought Chains for Science Question Answering},
  author = {Lu, Pan and Mishra, Swaroop and Xia, Tony and Qiu, Liang and Chang, Kai-Wei and Zhu, Song-Chun and Tafjord, Oyvind and Clark, Peter and Kalyan, Ashwin},
  booktitle = NeurIPS,
  year = {2022}
}

@inproceedings{GQA_CVPR19,
  title={Gqa: A new dataset for real-world visual reasoning and compositional question answering},
  author={Hudson, Drew A and Manning, Christopher D},
  booktitle=CVPR,
  pages={6700--6709},
  year={2019}
}

@inproceedings{VQAV2_CVPR17,
    author = {Yash Goyal and Tejas Khot and Douglas Summers{-}Stay and Dhruv Batra and Devi Parikh},
    title = {Making the {V} in {VQA} Matter: Elevating the Role of Image Understanding in {V}isual {Q}uestion {A}nswering},
    booktitle = CVPR,
    year = {2017},
}

@inproceedings{TextVQA_CVPR19,
  title={Towards vqa models that can read},
  author={Singh, Amanpreet and Natarajan, Vivek and Shah, Meet and Jiang, Yu and Chen, Xinlei and Batra, Dhruv and Parikh, Devi and Rohrbach, Marcus},
  booktitle=CVPR,
  pages={8317--8326},
  year={2019}
}

@inproceedings{masry-etal-2022-chartqa,
    title = "{C}hart{QA}: A Benchmark for Question Answering about Charts with Visual and Logical Reasoning",
    author = "Masry, Ahmed  and
      Long, Do  and
      Tan, Jia Qing  and
      Joty, Shafiq  and
      Hoque, Enamul",
    booktitle = ACL,
    year = "2022"
}

@article{huang2025keeping,
  title={Keeping yourself is important in downstream tuning multimodal large language model},
  author={Huang, Wenke and Liang, Jian and Guo, Xianda and Fang, Yiyang and Wan, Guancheng and Rong, Xuankun and Wen, Chi and Shi, Zekun and Li, Qingyun and Zhu, Didi and others},
  journal={arXiv preprint arXiv:2503.04543},
  year={2025}
}

@inproceedings{huang2025be,
  title={Be Confident: Uncovering Overfitting in MLLM Multi-Task Tuning},
  author={Huang, Wenke and Liang, Jian and Wan, Guancheng and Zhu, Didi and Li, He and Shao, Jiawei and Ye, Mang and Du, Bo and Tao, Dacheng},
  booktitle=ICML,
  year={2025}
}

@inproceedings{huang2024learn,
  title={Learn from Downstream and Be Yourself in Multimodal Large Language Model Fine-Tuning},
  author={Huang, Wenke and Liang, Jian and Shi, Zekun and Zhu, Didi and Wan, Guancheng and Li, He and Du, Bo and Tao, Dacheng and Ye, Mang},
  booktitle=ICML,
  year={2025}
}

@article{bi2024visual,
  title={Visual Instruction Tuning with 500x Fewer Parameters through Modality Linear Representation-Steering},
  author={Bi, Jinhe and Wang, Yujun and Chen, Haokun and Xiao, Xun and Hecker, Artur and Tresp, Volker and Ma, Yunpu},
  journal={arXiv preprint arXiv:2412.12359},
  year={2024}
}

@article{wei2022chain,
  title={Chain-of-thought prompting elicits reasoning in large language models},
  author={Wei, Jason and Wang, Xuezhi and Schuurmans, Dale and Bosma, Maarten and Xia, Fei and Chi, Ed and Le, Quoc V and Zhou, Denny and others},
  journal=NeurIPS,
  volume={35},
  pages={24824--24837},
  year={2022}
}

@article{touvron2023llama,
  title={Llama: Open and efficient foundation language models},
  author={Touvron, Hugo and Lavril, Thibaut and Izacard, Gautier and Martinet, Xavier and Lachaux, Marie-Anne and Lacroix, Timoth{\'e}e and Rozi{\`e}re, Baptiste and Goyal, Naman and Hambro, Eric and Azhar, Faisal and others},
  journal={arXiv preprint arXiv:2302.13971},
  year={2023}
}

@article{xing2024emo,
  title={Emo-llama: Enhancing facial emotion understanding with instruction tuning},
  author={Xing, Bohao and Yu, Zitong and Liu, Xin and Yuan, Kaishen and Ye, Qilang and Xie, Weicheng and Yue, Huanjing and Yang, Jingyu and K{\"a}lvi{\"a}inen, Heikki},
  journal={arXiv preprint arXiv:2408.11424},
  year={2024}
}

@article{yang2024emollm,
  title={Emollm: Multimodal emotional understanding meets large language models},
  author={Yang, Qu and Ye, Mang and Du, Bo},
  journal={arXiv preprint arXiv:2406.16442},
  year={2024}
}

@inproceedings{EmoVIT,
  title={Emovit: Revolutionizing emotion insights with visual instruction tuning},
  author={Xie, Hongxia and Peng, Chu-Jun and Tseng, Yu-Wen and Chen, Hung-Jen and Hsu, Chan-Feng and Shuai, Hong-Han and Cheng, Wen-Huang},
  booktitle=CVPR,
  pages={26596--26605},
  year={2024}
}

@inproceedings{zhaofacephi,
  title={FacePhi: Lightweight Multimodal Large Language Model for Facial Landmark Emotion Recognition},
  author={Zhao, Hongjin and Liu, Zheyuan and Liu, Yang and Qin, Zhenyue and Liu, Jiaxu and Gedeon, Tom},
  booktitle=ICLRW,
  year={2024}
}

@inproceedings{cheng2024emotion,
  title={Emotion-LLaMA: Multimodal Emotion Recognition and Reasoning with Instruction Tuning},
  author={Cheng, Zebang and Cheng, Zhi-Qi and He, Jun-Yan and Sun, Jingdong and Wang, Kai and Lin, Yuxiang and Lian, Zheng and Peng, Xiaojiang and Hauptmann, Alexander},
  booktitle=NeurIPS,
  year={2024}
}

@article{lian2025affectgpt,
  title={AffectGPT: A New Dataset, Model, and Benchmark for Emotion Understanding with Multimodal Large Language Models},
  author={Lian, Zheng and Chen, Haoyu and Chen, Lan and Sun, Haiyang and Sun, Licai and Ren, Yong and Cheng, Zebang and Liu, Bin and Liu, Rui and Peng, Xiaojiang and others},
  journal={arXiv preprint arXiv:2501.16566},
  year={2025}
}

@article{lian2023explainable,
  title={Explainable multimodal emotion reasoning},
  author={Lian, Zheng and Sun, Licai and Xu, Mingyu and Sun, Haiyang and Xu, Ke and Wen, Zhuofan and Chen, Shun and Liu, Bin and Tao, Jianhua},
  journal={CoRR},
  year={2023}
}

@article{lian2023explainable2,
  title={Explainable multimodal emotion recognition},
  author={Lian, Zheng and Sun, Haiyang and Sun, Licai and Gu, Hao and Wen, Zhuofan and Zhang, Siyuan and Chen, Shun and Xu, Mingyu and Xu, Ke and Chen, Kang and others},
  journal={arXiv preprint arXiv:2306.15401},
  year={2023}
}

@article{liu2025guardreasoner,
  title={GuardReasoner-VL: Safeguarding VLMs via Reinforced Reasoning},
  author={Liu, Yue and Zhai, Shengfang and Du, Mingzhe and Chen, Yulin and Cao, Tri and Gao, Hongcheng and Wang, Cheng and Li, Xinfeng and Wang, Kun and Fang, Junfeng and others},
  journal={arXiv preprint arXiv:2505.11049},
  year={2025}
}

@article{jin2025two,
  title={Two Heads are Better Than One: Test-time Scaling of Multi-agent Collaborative Reasoning},
  author={Jin, Can and Peng, Hongwu and Zhang, Qixin and Tang, Yujin and Metaxas, Dimitris N and Che, Tong},
  journal={arXiv preprint arXiv:2504.09772},
  year={2025}
}

@article{jin2024learning,
  title={Learning from teaching regularization: Generalizable correlations should be easy to imitate},
  author={Jin, Can and Che, Tong and Peng, Hongwu and Li, Yiyuan and Metaxas, Dimitris and Pavone, Marco},
  journal=NeurIPS,
  volume={37},
  pages={966--994},
  year={2024}
}

@article{wang2025safety,
  title={Safety in Large Reasoning Models: A Survey},
  author={Wang, Cheng and Liu, Yue and Li, Baolong and Zhang, Duzhen and Li, Zhongzhi and Fang, Junfeng},
  journal={arXiv preprint arXiv:2504.17704},
  year={2025}
}

@article{liu2024dora,
  title={Dora: Weight-decomposed low-rank adaptation},
  author={Liu, Shih-Yang and Wang, Chien-Yi and Yin, Hongxu and Molchanov, Pavlo and Wang, Yu-Chiang Frank and Cheng, Kwang-Ting and Chen, Min-Hung},
  journal={arXiv preprint arXiv:2402.09353},
  year={2024}
}

@article{zhao2024galore,
  title={Galore: Memory-efficient llm training by gradient low-rank projection},
  author={Zhao, Jiawei and Zhang, Zhenyu and Chen, Beidi and Wang, Zhangyang and Anandkumar, Anima and Tian, Yuandong},
  journal={arXiv preprint arXiv:2403.03507},
  year={2024}
}

@inproceedings{LoRASculpt_CVPR25,
    title={LoRASculpt: Sculpting LoRA for Harmonizing General and Specialized Knowledge in Multimodal Large Language Models},
    author={Liang, Jian and Huang, Wenke and  Wan, Guancheng and Yang, Qu and Ye, Mang},
    booktitle=CVPR,
    year={2025}
}

@inproceedings{GRPO_NeurIPS24,
  title={Group robust preference optimization in reward-free rlhf},
  author={Ramesh, Shyam Sundhar and Hu, Yifan and Chaimalas, Iason and Mehta, Viraj and Sessa, Pier Giuseppe and Bou Ammar, Haitham and Bogunovic, Ilija},
  booktitle=NeurIPS,
  pages={37100--37137},
  year={2024}
}

@article{SEEDGRPO_arXiv25,
  title={Seed-grpo: Semantic entropy enhanced grpo for uncertainty-aware policy optimization},
  author={Chen, Minghan and Chen, Guikun and Wang, Wenguan and Yang, Yi},
  journal={arXiv preprint arXiv:2505.12346},
  year={2025}
}

@article{DeepSeekR1_arXiv25,
  title={Deepseek-r1: Incentivizing reasoning capability in llms via reinforcement learning},
  author={Guo, Daya and Yang, Dejian and Zhang, Haowei and Song, Junxiao and Zhang, Ruoyu and Xu, Runxin and Zhu, Qihao and Ma, Shirong and Wang, Peiyi and Bi, Xiao and others},
  journal={arXiv preprint arXiv:2501.12948},
  year={2025}
}

@article{TreeRPO_arXiv25,
  title={TreeRPO: Tree Relative Policy Optimization},
  author={Yang, Zhicheng and Guo, Zhijiang and Huang, Yinya and Liang, Xiaodan and Wang, Yiwei and Tang, Jing},
  journal={arXiv preprint arXiv:2506.05183},
  year={2025}
}

@article{RLwithCoT_arXiv25,
  title={Delving into RL for Image Generation with CoT: A Study on DPO vs. GRPO},
  author={Tong, Chengzhuo and Guo, Ziyu and Zhang, Renrui and Shan, Wenyu and Wei, Xinyu and Xing, Zhenghao and Li, Hongsheng and Heng, Pheng-Ann},
  journal={arXiv preprint arXiv:2505.17017},
  year={2025}
}

@article{rajani2025scalpel,
  title={Scalpel vs. Hammer: GRPO Amplifies Existing Capabilities, SFT Replaces Them},
  author={Rajani, Neel and Gema, Aryo Pradipta and Goldfarb-Tarrant, Seraphina and Titov, Ivan},
  journal={arXiv preprint arXiv:2507.10616},
  year={2025}
}

@article{liu2025reinforcement,
  title={Reinforcement learning meets large language models: A survey of advancements and applications across the llm lifecycle},
  author={Liu, Keliang and Yang, Dingkang and Qian, Ziyun and Yin, Weijie and Wang, Yuchi and Li, Hongsheng and Liu, Jun and Zhai, Peng and Liu, Yang and Zhang, Lihua},
  journal={arXiv preprint arXiv:2509.16679},
  year={2025}
}

@article{zhou2025reinforced,
  title={Reinforced mllm: A survey on rl-based reasoning in multimodal large language models},
  author={Zhou, Guanghao and Qiu, Panjia and Chen, Cen and Wang, Jie and Yang, Zheming and Xu, Jian and Qiu, Minghui},
  journal={arXiv preprint arXiv:2504.21277},
  year={2025}
}

@article{xu2024dpo,
  title={Is dpo superior to ppo for llm alignment? a comprehensive study},
  author={Xu, Shusheng and Fu, Wei and Gao, Jiaxuan and Ye, Wenjie and Liu, Weilin and Mei, Zhiyu and Wang, Guangju and Yu, Chao and Wu, Yi},
  journal={arXiv preprint arXiv:2404.10719},
  year={2024}
}

@article{schulman2017proximal,
  title={Proximal policy optimization algorithms},
  author={Schulman, John and Wolski, Filip and Dhariwal, Prafulla and Radford, Alec and Klimov, Oleg},
  journal={arXiv preprint arXiv:1707.06347},
  year={2017}
}

@article{rafailov2023direct,
  title={Direct preference optimization: Your language model is secretly a reward model},
  author={Rafailov, Rafael and Sharma, Archit and Mitchell, Eric and Manning, Christopher D and Ermon, Stefano and Finn, Chelsea},
  journal=NeurIPS,
  volume={36},
  pages={53728--53741},
  year={2023}
}

@article{huang2025mapo,
  title={Mapo: Mixed advantage policy optimization},
  author={Huang, Wenke and Zhang, Quan and Fang, Yiyang and Liang, Jian and Rong, Xuankun and Yao, Huanjin and Wan, Guancheng and Liang, Ke and He, Wenwen and Li, Mingjun and others},
  journal={arXiv preprint arXiv:2509.18849},
  year={2025}
}

@article{zhang2025r1,
  title={R1-vl: Learning to reason with multimodal large language models via step-wise group relative policy optimization},
  author={Zhang, Jingyi and Huang, Jiaxing and Yao, Huanjin and Liu, Shunyu and Zhang, Xikun and Lu, Shijian and Tao, Dacheng},
  journal={arXiv preprint arXiv:2503.12937},
  year={2025}
}

@article{yao2025r1,
  title={R1-ShareVL: Incentivizing Reasoning Capability of Multimodal Large Language Models via Share-GRPO},
  author={Yao, Huanjin and Yin, Qixiang and Zhang, Jingyi and Yang, Min and Wang, Yibo and Wu, Wenhao and Su, Fei and Shen, Li and Qiu, Minghui and Tao, Dacheng and others},
  journal={arXiv preprint arXiv:2505.16673},
  year={2025}
}

@article{yu2025dapo,
  title={Dapo: An open-source llm reinforcement learning system at scale},
  author={Yu, Qiying and Zhang, Zheng and Zhu, Ruofei and Yuan, Yufeng and Zuo, Xiaochen and Yue, Yu and Dai, Weinan and Fan, Tiantian and Liu, Gaohong and Liu, Lingjun and others},
  journal={arXiv preprint arXiv:2503.14476},
  year={2025}
}

@article{zhao2025r1,
  title={R1-omni: Explainable omni-multimodal emotion recognition with reinforcement learning},
  author={Zhao, Jiaxing and Wei, Xihan and Bo, Liefeng},
  journal={arXiv preprint arXiv:2503.05379},
  year={2025}
}

@article{shao2024deepseekmath,
  title={Deepseekmath: Pushing the limits of mathematical reasoning in open language models},
  author={Shao, Zhihong and Wang, Peiyi and Zhu, Qihao and Xu, Runxin and Song, Junxiao and Bi, Xiao and Zhang, Haowei and Zhang, Mingchuan and Li, YK and Wu, Yang and others},
  journal={arXiv preprint arXiv:2402.03300},
  year={2024}
}

@article{robeyns2025improving,
  title={Improving LLM-Generated Code Quality with GRPO},
  author={Robeyns, Maxime and Aitchison, Laurence},
  journal={arXiv preprint arXiv:2506.02211},
  year={2025}
}

@article{zhang2024can,
  title={How can llm guide rl? a value-based approach},
  author={Zhang, Shenao and Zheng, Sirui and Ke, Shuqi and Liu, Zhihan and Jin, Wanxin and Yuan, Jianbo and Yang, Yingxiang and Yang, Hongxia and Wang, Zhaoran},
  journal={arXiv preprint arXiv:2402.16181},
  year={2024}
}

@article{gao2024designing,
  title={On designing effective rl reward at training time for llm reasoning},
  author={Gao, Jiaxuan and Xu, Shusheng and Ye, Wenjie and Liu, Weilin and He, Chuyi and Fu, Wei and Mei, Zhiyu and Wang, Guangju and Wu, Yi},
  journal={arXiv preprint arXiv:2410.15115},
  year={2024}
}

@article{huang2025boosting,
  title={Boosting mllm reasoning with text-debiased hint-grpo},
  author={Huang, Qihan and Dai, Weilong and Liu, Jinlong and He, Wanggui and Jiang, Hao and Song, Mingli and Chen, Jingyuan and Yao, Chang and Song, Jie},
  journal={arXiv preprint arXiv:2503.23905},
  year={2025}
}

@article{feng2025video,
  title={Video-r1: Reinforcing video reasoning in mllms},
  author={Feng, Kaituo and Gong, Kaixiong and Li, Bohao and Guo, Zonghao and Wang, Yibing and Peng, Tianshuo and Wu, Junfei and Zhang, Xiaoying and Wang, Benyou and Yue, Xiangyu},
  journal={arXiv preprint arXiv:2503.21776},
  year={2025}
}

@article{li2025videochat,
  title={Videochat-r1: Enhancing spatio-temporal perception via reinforcement fine-tuning},
  author={Li, Xinhao and Yan, Ziang and Meng, Desen and Dong, Lu and Zeng, Xiangyu and He, Yinan and Wang, Yali and Qiao, Yu and Wang, Yi and Wang, Limin},
  journal={arXiv preprint arXiv:2504.06958},
  year={2025}
}

@article{liu2025visual,
  title={Visual-rft: Visual reinforcement fine-tuning},
  author={Liu, Ziyu and Sun, Zeyi and Zang, Yuhang and Dong, Xiaoyi and Cao, Yuhang and Duan, Haodong and Lin, Dahua and Wang, Jiaqi},
  journal={arXiv preprint arXiv:2503.01785},
  year={2025}
}

@inproceedings{yang2023emoset,
  title={Emoset: A large-scale visual emotion dataset with rich attributes},
  author={Yang, Jingyuan and Huang, Qirui and Ding, Tingting and Lischinski, Dani and Cohen-Or, Danny and Huang, Hui},
  booktitle=ICCV,
  pages={20383--20394},
  year={2023}
}

@inproceedings{panda2018contemplating,
  title={Contemplating visual emotions: Understanding and overcoming dataset bias},
  author={Panda, Rameswar and Zhang, Jianming and Li, Haoxiang and Lee, Joon-Young and Lu, Xin and Roy-Chowdhury, Amit K},
  booktitle=ECCV,
  pages={579--595},
  year={2018}
}

@inproceedings{peng2015mixed,
  title={A mixed bag of emotions: Model, predict, and transfer emotion distributions},
  author={Peng, Kuan-Chuan and Chen, Tsuhan and Sadovnik, Amir and Gallagher, Andrew C},
  booktitle=CVPR,
  pages={860--868},
  year={2015}
}

@InProceedings{bai2025chat,
    author    = {Bai, Yang and Ji, Yucheng and Cao, Min and Wang, Jinqiao and Ye, Mang},
    title     = {Chat-based Person Retrieval via Dialogue-Refined Cross-Modal Alignment},
    booktitle = CVPR,
    year      = {2025}
}

@inproceedings{fang2025emoe,
  title     = {EMOE: Modality-Specific Enhanced Dynamic Emotion Experts},
  author    = {Fang, Yiyang and Huang, Wenke and Wan, Guancheng and Su, Kehua and Ye, Mang},
  booktitle = CVPR,
  year      = {2025}
}

@inproceedings{fang2025sepm,
  title     = {Catch Your Emotion: Sharpening Emotion Perception in Multimodal Large Language Models},
  author    = {Fang, Yiyang and Liang, Jian and Huang, Wenke and Li, He and Su, Kehua and Ye, Mang},
  booktitle = ICML,
  year      = {2025}
}

@article{rong2025backdoor,
  title={Backdoor Cleaning without External Guidance in MLLM Fine-tuning},
  author={Rong, Xuankun and Huang, Wenke and Liang, Jian and Bi, Jinhe and Xiao, Xun and Li, Yiming and Du, Bo and Ye, Mang},
  journal={arXiv preprint arXiv:2505.16916},
  year={2025}
}

@article{rong2025safegrpo,
  title={SafeGRPO: Self-Rewarded Multimodal Safety Alignment via Rule-Governed Policy Optimization},
  author={Rong, Xuankun and Huang, Wenke and Wang, Tingfeng and Zhou, Daiguo and Du, Bo and Ye, Mang},
  journal={arXiv preprint arXiv:2511.12982},
  year={2025}
}

@article{SafetySurvey,
  title={A survey of safety on large vision-language models: Attacks, defenses and evaluations},
  author={Ye, Mang and Rong, Xuankun and Huang, Wenke and Du, Bo and Yu, Nenghai and Tao, Dacheng},
  journal={arXiv preprint arXiv:2502.14881},
  year={2025}
}
}


\end{document}